\documentclass[letterpaper]{article}

\usepackage{aaai20}
\usepackage{times}
\usepackage{helvet}
\usepackage{courier}
\usepackage{hyperref}
\usepackage{url}
\usepackage{graphicx}
\usepackage{epsfig}
\usepackage{amsmath}
\usepackage{amssymb}
\usepackage{algorithm}
\usepackage{algorithmic}
\usepackage{booktabs}
\usepackage{dblfloatfix}
\usepackage{enumitem}

\def\rvx{{\mathbf{x}}}
\def\rvz{{\mathbf{z}}}
\def\sR{{\mathbb{R}}}

\title{Variational Inference with Latent Space Quantization for Adversarial Resilience}
\author{Vinay Kyatham\thanks{Equal contribution}\\SigTuple Technologies\\Bangalore, India \And Mayank Mishra\footnotemark[1], Tarun Kumar Yadav, \\ \textbf{\Large Deepak Mishra\thanks{Equal contribution}, Prathosh AP\footnotemark[2]}\\Indian Institute of Technology Delhi\\New Delhi, India}

\begin{document}
\maketitle

\begin{abstract}
    Despite their tremendous success in modelling high-dimensional data manifolds, deep neural networks suffer from the threat of adversarial attacks - Existence of perceptually valid input-like samples obtained through careful perturbation that lead to degradation in the performance of the underlying model. Major concerns with existing defense mechanisms include non-generalizability across different attacks, models and large inference time. In this paper, we propose a generalized defense mechanism capitalizing on the expressive power of regularized latent space based generative models. We design an adversarial filter, devoid of access to classifier and  adversaries, which makes it usable in tandem with any classifier. The basic idea is to learn a Lipschitz constrained mapping from the data manifold, incorporating adversarial perturbations, to a quantized latent space and re-map it to the true data manifold. Specifically, we simultaneously auto-encode the data manifold and its perturbations implicitly through the perturbations of the regularized and quantized generative latent space, realized using variational inference. We demonstrate the efficacy of the proposed formulation in providing resilience against multiple attack types (black and white box) and methods, while being almost real-time. Our experiments show that the proposed method surpasses the state-of-the-art techniques in several cases.
\end{abstract}

\section{Introduction}

Deep neural networks have shown tremendous success in various computer vision tasks. One of the primary factors contributing to their success is the availability of abundant data. This generally leads to an incomplete exploration of data space with the available training set, which in-turn results in loopholes in the data manifold \cite{gilmer2018adversarial,schmidt2018adversarially}. Adversarial attacks exploit these gaps in the data manifold, unexplored by the classifier which leads to the failure of otherwise successful networks. This unexplored subspace, called \textit{adversarial subspace}, contains adversarial samples generated using perturbation of original training samples with carefully designed synthetic noise \cite{goodfellow2014explaining,szegedy2013intriguing,moosavi2016deepfool,carlini2017towards,madry2017towards}. This is an important concern not only from a point of security but also from a generalization perspective \cite{tsipras2018robustness}. 
In the rest of the section, we will present an overview of the existing adversarial attacks, defense mechanisms along with the motivation for our work.

\subsection{Adversarial attacks - General principles}
An adversarial sample is obtained by perturbing the input sample with a small amount such that its perceptual quality is unaltered but the class label is changed under the classifier. Formally, let $\rvx \in \mathbb{R}^N$ denote a sample from the natural data manifold and $\rvx' =\rvx + \delta $, denote a perturbation on $\rvx$ with $\delta \in \mathcal{S}$, $\mathcal{S} \subseteq \mathbb{R}^N $ being the space of all possible perturbations. Under a given distance metric $\mathcal{D}$ and a classification scheme $h(\rvx)$, the sample $\rvx'$ is called an adversarial example for $\rvx$ if  $\mathcal{D(\rvx,\rvx')} \leq \epsilon$ and $h(\rvx) \neq h(\rvx') $. A large body of attacks consider a $l_p$-norm based $\mathcal{D}$, with $l_2$ and $l_\infty$ norms being the significant ones, and solve an optimization problem on the loss function of $h(\rvx)$ to obtain the desired $\delta$.
Attacks can be targeted so that the classifier is misguided to a specific class or non-targeted so that it outputs an arbitrary class different from the original class. Further categorization of adversarial attacks is based on the level of access the attacker has about the classification and defense scheme. Specifically they are defined as white and black box attacks.


\subsection{Attack methods}

There is a gamut of literature on creating adversarial attacks. \citeauthor{goodfellow2014explaining} propose Fast Gradient Sign Method (FGSM) that performs a one step gradient update along the direction of sign of gradient of loss at each pixel. \citeauthor{kurakin2016adversarial} introduced Basic Iterative method (BIM) which runs FGSM for a few iterations. Deepfool \cite{moosavi2016deepfool} is another iterative attack which computes adversarial perturbations through an orthogonal projection of the sample on the decision boundary. Carlini-Wagner (CW) attack \cite{carlini2017towards} is an optimization based attack that uses logits-based objective function instead of the commonly used cross-entropy loss. We choose these attacks since they cover a good breadth of the class of attacks.

\section{Prior art on defense mechanisms}

A large number of defense mechanisms to diminish the effect of adversarial attacks are available \cite{goodfellow2016deep,samangouei2018defensegan,ilyas2017robust,song2017pixeldefend,shen2017ape,meng2017magnet,madry2017towards,papernot2016distillation}. Broadly, these can be divided into the following categories -


\begin{enumerate}
    \item \textbf{Adversarial retraining} - A natural way to make the classifier robust against the adversaries is to retrain the classifier using adversarial examples \cite{goodfellow2016deep,szegedy2013intriguing}. Several improvisations of adversarial retraining have been proposed \cite{madry2017towards,sinha2017certifying,tramer2017ensemble}. While adversarial retraining is a simple method for defense and is robust to first-order adversaries, it is shown to be ineffective towards DeepFool and CW attack \cite{tramer2017ensemble} and also black-box attacks \cite{2014arXiv1412.6572G,sharma2017attacking,ding2019sensitivity}
    \item \textbf{Modified training} - Here the idea is to tweak the training procedure and/or the training examples of the classifier so that the decision boundary learnt is robust to adversarial examples \cite{papernot2016distillation,guo2017countering,xiao2018training,dhillon2018stochastic,xie2017mitigating}. However, \citeauthor{athalye2018obfuscated} demonstrate that most of these defenses are vulnerable because they capitalize on obfuscated gradients that can be mitigated.
    \item \textbf{Adversarial filtering} - These defense mechanisms pre-process the adversarial examples to make them non-adversarial either by manifold projection or by using generative models. For example, MagNet \cite{meng2017magnet} trains a collection of detector networks that differentiate between normal and adversarial examples. It also includes a reformer network (one or a collection of auto-encoders) to push adversarial examples close to the data manifold. A recent strategy called Defense-GAN \cite{samangouei2018defensegan} trains a generative adversarial network (GAN) \cite{goodfellow2014generative} only on legitimate examples and uses it to denoise adversarial examples. At the time of inference, they find images from the range of generator that are near the input image but lie on the legitimate data manifold. 
    This requires $L$ iterations of back propagation for $R$ random initializations to find the nearest legitimate image, typical value of L is 200 and R is 10. Other GAN based defences, for example, PixelDefend \cite{song2017pixeldefend} and APE-GAN \cite{shen2017ape}, perform image-to-image translation to convert an adversarial image into a legitimate image.
\end{enumerate}

\begin{figure*}[t!]
    \centering
    \includegraphics[width=0.9 \textwidth]{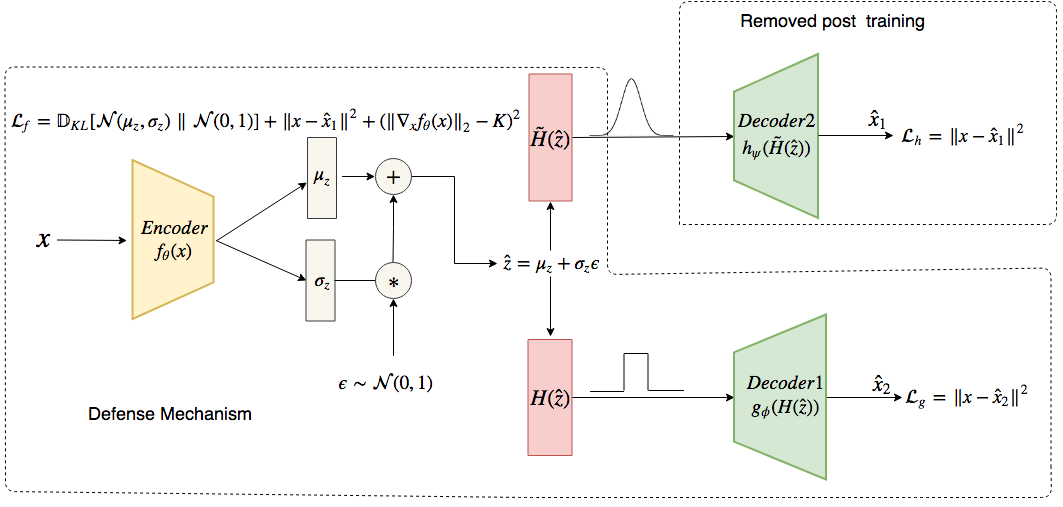}
    \caption{Proposed LQ-VAE - A Lipschitz constrained encoder ($L_f$) encodes the input image into a latent space quantized by the function $H$ which is explored through a stochastic perturbation ($\epsilon$). During inference, Decoder1 maps the quantized latent codes generated by the adversarial images back to the image space. Training is done \textbf{only on real data samples} by using an approximate differentiable version of Decoder1 (i.e. Decoder2).}
    \label{lqvae}
\end{figure*}

\subsection{Problem setting and our contributions}

As mentioned, while the existing defense mechanisms have their own merits, each of them suffer from aforementioned disadvantages. 
\citeauthor{athalye2018obfuscated} argue that most of these methods give a false sense of security since most of them capitalize on masking the gradients (obfuscated gradients) so that it is difficult to generate adversarial examples. However, they show that this can be circumvented using techniques such as approximation of derivatives by a differentiable approximation of the function, reparameterization and computation of expectations. In this paper, we propose a defense mechanism based on quantized latent variable generative autoencoders to alleviate the aforementioned issues. Our contributions are enlisted below:

\begin{enumerate}
    \item Propose a quantized latent variable generative model based defense mechanism devoid of access to classifier and adversaries.
    \item Construction of a Lipschitz constrained latent encoder that preserves the distances under a metric space on the latent and the data manifolds.
    \item Constraining the latent space to follow a known distribution so that stochastic exploration of the latent neighbourhood corresponding to the data neighbourhood is possible.
    \item Use of dual decoders with a quantization on the latent space so that a large neighbourhood around a data sample is explored and easily remapped back to the data point.
\end{enumerate}

    
    
    
    
    

\section{Proposed Method}

\subsection{Motivation}


In many previous works it is hypothesized that the adversarial examples fall off the data manifold \cite{lee2017generative,samangouei2018defensegan}. This suggests that a defense model could potentially be built by replacing an adversarial example with the nearest correctly-classified sample from the data manifold. However, searching in high-dimensional data manifolds is expensive, not generalizable and moreover, it has been found that the adversarial examples might fall on the data manifold too \cite{gilmer2018adversarial}. Thus, a better approach could be to project the data manifold onto an explorable compact generative latent space and remap the latent codes back to the legitimate data. If the latent space projector is made to be Lipschitz constrained and compact, then one can hope that the adversarial examples adhere to a latent code that is invertible to the legitimate data.

\subsection{Lipschitz constrained latent transformation}

Let $f(\rvx): \sR^N \rightarrow \sR^M$ be a Lipschitz constrained function which maps a data point $\rvx$ of dimension $N$ to a latent vector $\rvz$ of dimension $M$ such that $M < N$. For such an $f$, if the adversarial perturbation $\delta$ on $\rvx$ is bounded then the equivalent perturbation $\delta_z$ on $\rvz$ is also bounded.




Since the goal is to learn mappings in high-dimensional spaces, we use Deep Neural Networks (DNNs) parameterized by $\theta$, to approximate $f_\theta$. There have been many methods proposed to make a DNN $K$-Lipschitz including gradient clipping \cite{arjovsky2017wasserstein} and gradient norm penalty \cite{gulrajani2017improved}. We employ gradient norm penalty on the encoder since it is observed to be more stable.




\subsection{Latent exploration via variational inference}

\begin{figure}[H]
    \includegraphics[width=0.4\textwidth]{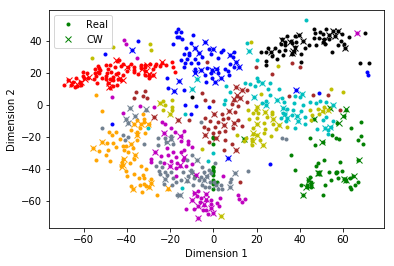}
    \caption{t-SNE plot of the latent encodings of mnist real images and CW adversarial images. It can seen that the embeddings of real and adversarial data overlap.} 
    \label{fig:t-SNE plot}
\end{figure}

As mentioned earlier, the goal is to explore the latent neighbourhood induced by perturbing a given input sample. This effectively means that one has to sample from the true conditional distribution $p(\hat{\rvz}|\rvx)$. However since there is no direct access to $p(\hat{\rvz}|\rvx)$ we propose to use the principles of variational inference \cite{kingma2013auto}, where sampling from $p(\hat{\rvz}|\rvx)$ is facilitated by approximating it with a variational distribution $q_\theta(\hat{\rvz}|\rvx)$ on $\hat{\rvz}$ that is parameterized by the encoder network $f_\theta$. Now minimizing the KL-divergence between $p(\hat{\rvz}|\rvx)$ and $q_\theta(\hat{\rvz}|\rvx)$ results in the maximization of the so-called evidence lower bound given as follows:
\begin{equation}
    \mathcal{L=\mathbb{E}}_{q_{\theta}(\hat{\rvz}|\rvx)}[\log p_{\phi}(\rvx|\hat{\rvz})]-\mathbb{D}_{KL}\left[q_{\theta}(\hat{\rvz}|\rvx)||p_{\theta}(\hat{\rvz})\right]
    \label{elbo}
\end{equation}
where $p_{\phi}(\rvx | \hat{\rvz})$ represents a probabilistic decoder network that maps the latent space back to the data space and $p_{\theta}(\hat{\rvz})$ is an arbitrary prior on $\hat{\rvz}$ which is usually a Normal distribution. We propose to sample $\hat{\rvz} $ from  $q_{\theta}(\hat{\rvz}|\rvx)$  using the encoder network $f_\theta$ through the reparameterization trick \cite{kingma2013auto}. Thus, given a true data example, a cloud of perturbations is created around its latent representation obtained through Lipschitz constrained encoder $f_\theta$,  via variational sampling. The probabilistic decoder $p_{\phi}(\rvx | \hat{\rvz})$, parameterized using a neural network $g_\phi$, is then tasked to map all the points within that cloud to a single input example through maximization of the likelihood term in \eqref{elbo}, as shown in Figure \ref{lqvae}. Mathematically, if the encoder embeds and $\rvx$ to a $\hat{\rvz}$ and $\rvx + \delta$ to a $\hat{\rvz}'$, such that $||\hat{\rvz} - \hat{\rvz}'|| \leq |\delta_z|$ then decoder learns $g_\phi(\hat{\rvz}) = g_\phi(\hat{\rvz}') = \rvx$. During inference, when the encoder is presented with an adversarial example, it will place it within the learned latent cloud so that the decoder converts it into a non-adversarial sample. This fact has been illustrated in Figure \ref{fig:t-SNE plot} where a 2D t-SNE plot of the latent encodings (from the Lipschitz constrained encoder) of the true and the CW $l_2$ attacked adversarial samples from the MNIST data is shown. It can be seen that embeddings of the adversaries are extremely close to those of the true samples.




\subsection{Latent space quantization}

A recent work by \citeauthor{gilmer2018adversarial} makes a very important observation - closeness from a correctly classified sample does not guarantee non-adversarial nature. In other words, since the latent space is real-valued it is impossible to explore it in its entirety. Thus, there is a high chance that a probabilistic decoder $g_\phi$ is unable to remove the adversarial noise even when the latent vector for an adversarial example falls within the seen latent cloud.  We propose to address this issue by quantizing the latent space before it is input to the decoder.  Specifically, we design a fixed discrete quantization function  $H$ applied on each dimension of the real-output of the encoder as follows:
\begin{equation}
    \rvz_i = H_i(\hat{\rvz}) = \left\{
        \begin{array}{ll}
            +1 & \text{if \hspace{3mm}} |\hat{\rvz}_i| \le \eta
            \\
            -1 & \text{otherwise}
        \end{array}
        \right.
\end{equation}
where $\eta$ is the quantization threshold. $H(.)$ thus converts $\hat{\rvz}$ into binary coded vectors $\rvz$ thereby ensuring that the decoder $g_\phi$ receives a single latent code for all the input samples that map within a small neighbourhood of $\hat{\rvz}$. In contrast to real $\hat{\rvz}$, where $g_\phi$ has to learn a non-injective mapping, quantization allows it to learn a mapping close to injective since the same code vector is produced for all $\hat{\rvz}$ in this neighbourhood, hence making the training easier. This procedure potentially increases the robustness of the model too since the goal of an adversarial resilience model is not to exactly reproduce the non-adversarial version of a given sample but to produce an approximate version that is non-adversarial. Thus, it is imperative to just look for the presence or absence of the salient features that preserve the identity of a given example, which is accomplished by the binary quantization with the threshold $\eta$ being a hyperparameter whose value is chosen as, however not limited to, $0.67449$ since it gives equal probability to a bit being +1 or -1. Thus, any deviation in the input sample falling outside the latent cloud leads to flipping of the bits in the quantized space.

\subsubsection{Theorem 1}\label{theorem1}Let $\hat{\rvz}$ and $\hat{\rvz}_{adv}$ be the latent encoding of $\rvx$ and $\rvx_{adv}$ respectively. Let $\rvz = H(\hat{\rvz})$ and $\rvz_{adv} = H(\hat{\rvz}_{adv})$ be the corresponding quantizations. Then the probability of a particular bit being flipped is given by
\begin{equation}
    p = \int_{\eta - |\delta_z|}^{\eta + |\delta_z|} p_{\hat{\rvz}} (\hat{\rvz}) d\hat{\rvz}
\end{equation}
and the probability of $k = 6$ ($\approx 10\%$) bit flips for $M = 64$ dimensional latent space is ${M \choose k} p^k (1-p)^{M-k} = 1.177 \times 10^{-3}$ which is significantly low, when $\eta = 0.67449$ is chosen such that both bits are equally likely, $|\delta_z| = K|\delta| = 0.1 \times 0.3 = 0.03$ and $p_{\hat{\rvz}} (\hat{\rvz}) = \mathcal{N}(0, 1)$ (proof in supplementary material).

Figure \ref{fig:bit} depicts the bit-flippings in the latent codes of the CW adversaries on the MNIST data - It can be seen that about 90 \% of the total adversaries undergo less than 6\% of bits being flipped resulting in high classification accuracy, confirming the effectiveness of the decoder in ignoring the bit-flippings. Further, the binary encoding layer ensures that gradient produced at that layer is either 0 or undefined, thereby making a gradient based attack on the defense mechanism impossible.

\begin{figure}[t!]
    \hspace*{0.5cm} 
    \includegraphics[width=0.4\textwidth]{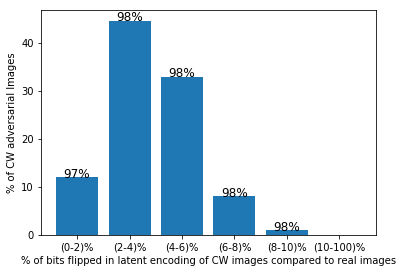}
    \caption{Effect of binary quantization on adversaries - It can be seen that more than 90 \% of the adversaries have less than 6\% of the bit flippings with high classification accuracy.} 
    \label{fig:bit}
\end{figure}
  
  
\subsection{Adversarial resilience by LQ-VAE}

As mentioned above, quantization prevents the flow of gradients through the encoder network $f_\theta$ that makes it non-trainable along with the decoder $g_\phi$. We, therefore, create a copy of $g_\phi$, say $h_\psi$, which uses soft quantization in place of $H$ (called $\tilde{H}$) and enable the training of $f_\theta$. A complete overview of the proposed method, called as Lisphitz-constrained quantized variational auto-encoder (LQ-VAE) is shown in Figure 1. Algorithm \ref{alg:algorithm} gives the details of the LQ-VAE training procedure where the likelihood term in \eqref{elbo} is approximated using mean-squared error term between the input and the output.  Note that the network $h_\psi$ is only used for training, not for inference. The defense scheme only contains the pipeline of $f_\theta-H-g_\phi$.

\begin{algorithm}[t!]
    \caption{LQ-VAE algorithm}
    \label{alg:algorithm}
    
    \hspace*{\algorithmicindent} \textbf{Input}: Dataset $\mathcal{D}$, Batchsize $B$, Encoder $f_{\theta}$, Decoder1  \hspace*{1cm} $g_{\phi}$, Learning rate $\eta$, Quantization functions $H$, $\Tilde{H}$ \\
    \hspace*{\algorithmicindent} \textbf{Output} Parameters $\theta^*$, $\phi^*$
    
    \begin{algorithmic}[1] 
    \STATE Make a copy $h_{\psi}$ of decoder $g_{\phi}$
    \REPEAT
    \STATE Sample $\{\rvx^{(1)} \cdots \rvx^{(B)} \}$ from dataset $\mathcal{D}$
    \STATE $\mu_\rvz^{(i)}, \sigma_\rvz^{(i)} \gets f_{\theta} ( \rvx^{(i)} )$
    \STATE Sample $\hat{\rvz}^{(i)}$ from $\mathcal{N}$($\mu_\rvz^{(i)}$,\,${\sigma_\rvz^{(i)}}^2$)
    \STATE $\hat{\rvx}_{1}^{(i)} \gets h_{\psi}(\tilde{H}(\hat{\rvz}^{(i)}))$
    \STATE $\hat{\rvx}_{2}^{(i)} \gets g_{\phi}(H(\hat{\rvz}^{(i)}))$
    \STATE $ \mathcal{L}_{h} \gets \sum_{i=1}^{B} \left\Vert \rvx^{(i)} - \hat{\rvx}_{1}^{(i)}\right\Vert_2^2 $
    \STATE $ \mathcal{L}_{g} \gets \sum_{i=1}^{B} \left\Vert \rvx^{(i)} - \hat{\rvx}_{2}^{(i)}\right\Vert_2^2 $
    \STATE $ \mathcal{L}_{f} \gets \mathcal{L}_{h} + \sum_{i=1}^{B} \mathbb{D}_{KL} \left[ \mathcal{N} ( \mu_\rvz^{(i)},{\sigma_\rvz^{(i)}}^2 ) || \mathcal{N}(0, 1) \right] + \hspace*{1cm} \sum_{i=1}^{B} \left[ \left\Vert \nabla_x f_{\theta}(x^{(i)})\right\Vert_2 - K\right]^2 $
    \STATE $ \theta \gets \theta + \eta \nabla_{\theta}\mathcal{L}_{f} $
    \STATE $ \phi \gets \phi + \eta \nabla_{\phi}\mathcal{L}_{g} $
    \STATE $ \psi \gets \psi + \eta \nabla_{\psi}\mathcal{L}_{h} $
    \UNTIL{ convergence of $\theta$, $\phi$ }
    \end{algorithmic}
\end{algorithm}

\section{Experiments and results}
We consider MNIST \cite{lecun2010mnist}, FMNIST\footnote{\url{https://github.com/zalandoresearch/fashion-mnist}}, and CelebA \cite{liu2015deep} datasets and use three classifier architectures, named A, B, and C, for black box and white box experiments\footnote{Architectures of all the classifiers and LQ-VAE is included in supplementary material.}. For MNIST and FMNIST, the standard 10 class classification task is considered, whereas for CelebA, a binary classification task of gender classification is taken, with accuracy as the metric. Five attack types namely, FGSM with $\epsilon=0.3$ \cite{goodfellow2014explaining}, $l_2$ CW \cite{carlini2017towards}, Deepfool \cite{moosavi2016deepfool}, iterative FGSM \cite{kurakin2016adversarial} and Madry \cite{madry2017towards}, generated from cleverhans library \cite{papernot2016cleverhans}, are considered for experimentation as they cover a good breadth of attack types. We compare our results with three defense strategies - Defense GAN \cite{samangouei2018defensegan}, Madry \cite{madry2017towards} and Adversarial retraining \cite{goodfellow2014explaining} based on the following facts (i) Defense GAN is close to our work in spirit. Their method also employs a generative model (GAN) and does not train on adversarial examples, (ii) Madry retraining is claimed to be a robust defense against all first-order gradient computation based attacks and (iii) adversarial retraining is one of the earliest benchmark defenses created. We use the Adam optimizer ($\beta_1 = 0.9, \beta_2 = 0.99$) with a learning rate of $10^{-3}$ to train LQ-VAE. We study different attack types, namely black box attacks, white box attack on classifier and end-to-end white box attack.

\begin{enumerate}
    \item{\textbf{Black box attacks} - The attacker neither has access to the original classifier nor the defense scheme, rather he has to generate adversarial examples on a substitute classifier \cite{papernot2017practical}.}
    \item{\textbf{White box attacks} - We subdivide white box attacks into two further categories:}
    \begin{enumerate}
        \item \textbf{White box attacks on classifier} - In this case, it is assumed that the attacker has access to the original classifier but not the defense scheme and the adversarial examples are generated by computing gradients over the original classifier.
        \item \textbf{White box attacks with BPDA} - The attacker has access to both the original classifier and the defense mechanism. In other words, the adversarial examples can be generated by computing gradients over the LQ-VAE - classifier combination. \citeauthor{athalye2018obfuscated} argue that such methods can be attacked using Backward Pass Differentiable Approximation (BPDA). We attack the defense mechanism by approximating the discretization function $H(\hat{\rvz})$ with $\tilde{H}(\hat{\rvz})$ given as follows:
        \begin{equation}
            \tilde{H}_i(\hat{\rvz}) = \frac{\eta^2 - \hat{\rvz}_i^2}{c + |\eta^2 - \hat{\rvz}_i^2|}
        \end{equation}
        where $c$ is a non-negative hyperparameter. It can be seen that $H(\hat{\rvz}) = \tilde{H}(\hat{\rvz})$ when $c = 0$.
    \end{enumerate}
\end{enumerate}


\subsection{Results on white box attacks on classifier}
For MNIST and FMNIST, we use 60,000 real images for training the defense mechanisms and 10,000 adversarial images on the standard test set for testing. For CelebA, we use 90\% images for training and remaining for testing. Classification accuracy of all three classifier models, A, B, and C, in presence and absence of defense mechanisms are listed in Table 1, 2, and 3 for MNIST, FMNIST, and CelebA, respectively. The attacks reduce classification accuracy of all the models drastically. Adversarial retraining is able to defend FGSM attack up to a certain extent but fails on the other attacks, since the retraining is performed using adversarial examples generated by FGSM attacks. Similarly Madry, which also uses first order gradients based adversarial images to retrain the classifiers, shows consistent performance against FGSM. However, for Deepfool and CW attack, its performance is lower than Defense-GAN and the proposed LQ-VAE. Both, Defense-GAN and LQ-VAE, do not require adversarial augmentation, however, we note that LQ-VAE consistently outperforms Defense-GAN.

\subsection{Results on white box attacks with BPDA}
Most of the defense mechanisms that rely on gradient obfuscation \cite{athalye2018obfuscated} are broken completely by attacking using BPDA. However, Defense-GAN still has 55\% accuracy on MNIST after the attack since it does not rely on gradient obfuscation. Our work is similar in spirit to Defense-GAN in thse sense that it also does not rely on gradient obfuscation. We generate adversarial examples for the LQ-VAE - classifier combination via BPDA, however, passing these examples to the Lipschitz constrained encoder still results in these adversarial examples getting the same discrete latent codes (refer Figure \ref{fig:bit} and \nameref{theorem1}) as their non-adversarial counterparts and thus their non-adversarial counterparts are successfully reconstructed by the decoder resulting in their correct classification (refer Table \ref{tab:BPDA}). We compare our results with Defense-GAN since \citeauthor{athalye2018obfuscated} show that it can also be attacked using BPDA.


\begin{table*}[!t]
\begin{center}
\begin{tabular}{ccccccccc}
\hline
\multicolumn{1}{c}{\bf Attack}
&\multicolumn{1}{c}{\bf Model}
&\multicolumn{1}{c}{\bf No Attack}
&\multicolumn{1}{c}{\bf No Defense}
&\multicolumn{1}{c}{\bf LQ-VAE}
&\multicolumn{1}{c}{\bf Defense-GAN}
&\multicolumn{1}{c}{\bf Madry}
&\multicolumn{1}{c}{\bf Adv Tr}
\\ \hline
{FGSM}&A & 92.76 & 11.50 & 77.00 & 69.75 & 78.87 & 53.72\\
    &B & 91.17 & 10.14 & 69.41  & 56.72 & 76.94 & 59.79\\
    &C & 89.06 & 11.60 & 67.07 & 56.34 & 64.16 & 66.43\\
    \hline
{Deepfool}&A & 92.76 & 5.29 & 79.30 & 77.48 & 57.17 & 6.52\\
    &B & 91.17 & 6.54 & 79.41 & 74.97 & 52.58 & 14.74\\
    &C & 89.06 & 7.65 & 79.89 & 74.82 & 39.93 & 24.71\\
    \hline
{CW}&A & 92.76 & 5.41 & 80.64 & 78.75 & 62.55 & 5.35\\
    &B & 91.17 & 6.61 & 81.58 & 78.18 & 56.48 & 6.35\\
    &C & 89.06 & 7.89 & 82.31 & 78.58 & 43.72 & 8.00\\
    \hline
{\bf Average}&& 91.00 & 8.07 & \bf 77.40 & 71.73 & 59.15 & 27.29\\
\hline
\end{tabular}
\caption{Classification accuracy of the FMNIST classifiers on white box attacks with various defense strategies}
\end{center}
\label{tab:fmnistW}
\end{table*}

\begin{table*}[!t]
\begin{center}
\begin{tabular}{cccccccc}
\hline
\multicolumn{1}{c}{\bf Attack}
&\multicolumn{1}{c}{\bf Model}
&\multicolumn{1}{c}{\bf No Attack}
&\multicolumn{1}{c}{\bf No Defense}
&\multicolumn{1}{c}{\bf LQ-VAE}
&\multicolumn{1}{c}{\bf Defense-GAN}
&\multicolumn{1}{c}{\bf Madry}
&\multicolumn{1}{c}{\bf Adv Tr}
\\ \hline
{FGSM}&A & 99.40 & 20.16 &89.17 & 90.43 & 96.85 & 67.95\\
    &B & 99.41 & 13.17 &86.70 & 88.52 & 96.20 & 49.49\\
    &C & 98.37 & 5.66 &83.02  & 86.7 & 84.71 & 80.75\\
    \hline
{Deepfool}&A & 99.40 & 7.38 & 97.60 & 95.41 & 67.82 & 3.10\\
    &B & 99.41 & 5.88 & 97.74 & 93.03 & 66.35 & 5.75\\
    &C & 98.37 & 48.24 & 97.42 & 92.32 & 62.38 & 10.97\\
    \hline
{CW}&A & 99.40 & 8.85 & 97.66 & 94.37 & 69.15 & 1.20\\
    &B & 99.41 & 5.07 & 97.20 & 90.56 & 71.35 & 1.45\\
    &C & 98.37 & 8.44 & 97.36  & 92.5 & 58.65 & 2.15\\
    \hline
{\bf Average}&& 99.06 & 13.65 & \bf 93.76 & 91.54 & 74.83 & 24.76\\
\hline
\end{tabular}
\caption{Classification accuracy of the MNIST classifiers on white box attacks with various defense strategies.}
\end{center}
\label{tab:mnistW}
\end{table*}
 
\begin{table*}[!t]
\begin{center}
\begin{tabular}{ccccccccc}
\hline
\multicolumn{1}{c}{\bf Attack}
&\multicolumn{1}{c}{\bf Model}
&\multicolumn{1}{c}{\bf No Attack}
&\multicolumn{1}{c}{\bf No Defense}
&\multicolumn{1}{c}{\bf LQ-VAE}
&\multicolumn{1}{c}{\bf Defense-GAN}
&\multicolumn{1}{c}{\bf Madry}
&\multicolumn{1}{c}{\bf Adv Tr}
\\ \hline
{FGSM}&A & 96.34 & 3.65 & 81.04& 74.13 & 62.35 & 4.53\\
    &B & 96.60 & 3.40 & 64.74  & 67.06 & 71.42 & 72.88\\
    &C & 95.02 & 28.62 & 61.48 & 53.76 & 61.35 & 42.55\\
\hline
{Deepfool}&A & 96.34 & 3.56 & 85.89  & 83.87 & 52.86 & 6.26\\
    &B & 96.60 & 2.43 & 83.81  & 83.65 & 49.39 & 14.17\\
    &C & 95.02 & 10.92 & 62.79 & 78.56 & 42.37 & 38.45\\
\hline
{CW}&A & 96.34 & 6.98 & 85.90 & 84.64 & 58.62 & 11.88\\
    &B & 96.60 & 6.88 & 86.29 & 86.01 & 60.33 & 12.91\\
    &C & 95.02 & 10.92 & 79.20 & 78.56 & 45.02 & 38.45\\
\hline
{Iter FGSM}&A & 96.34 & 3.12 & 85.44 & 81.00 & 82.34 & 3.50\\
    &B & 96.60 & 3.55 & 72.29 & 72.05 & 72.19 & 9.16\\
    &C & 95.02 & 11.92 & 52.12 & 42.13 & 90.87 & 19.47\\
\hline
{Madry}&A & 96.34 & 2.84 & 85.11  & 81.43 & 76.35 & 3.52\\
    &B & 96.60 & 3.12  &70.01 & 74.01 & 70.32 & 8.52\\
    &C & 95.02 & 8.57 &54.00 & 45.11 & 84.09 & 18.59\\
    \hline
{\bf Average}&& 95.99 & 7.37 & \bf 74.01 & 72.40 & 65.32 & 20.32\\
\hline
\end{tabular}
\caption{Classification accuracy of the CelebA classifiers on white box attacks with various defense strategies.}
\end{center}
\label{tab:celebW}
\end{table*}

\begin{table*}[!t]
\begin{center}
\begin{tabular}{ccccccc}
\hline
{\bf Classifier}
&{\bf No Attack}
&{\bf No Defense}
&{\bf LQ-VAE}
&{\bf Defense-GAN}
&{\bf Madry}
&{\bf Adv Tr}\\
{\bf Substitute}&&&&&&
\\ \hline 
A/B & 99.40 & 33.32 & 88.09 & 89.14 & 97.17 & 95.78 \\
A/C & 99.40 & 45.35 & 90.60 & 90.08 & 98.27 & 96.82 \\
B/A & 99.41 & 42.22 & 90.63 & 91.40 & 97.38 & 94.64 \\
B/C & 99.41 & 38.73 & 89.92 & 89.89 & 98.03 & 95.30 \\
C/A & 98.37 & 28.93 & 91.98 & 90.90 & 90.59 & 32.12 \\
C/B & 98.37 & 18.01 & 89.38 & 88.73 &  89.14 & 21.79 \\
\hline
{\bf Average}& 99.06 & 34.43 & 90.10 & 90.02 & \bf 95.10 & 72.74\\
\hline
\end{tabular}
\caption{Classification accuracy of the MNIST classifier on FGSM black box attack images generated using substitute model.}
\end{center}
\label{tab:mb} 

\end{table*}
\begin{table*}[!t]
\label{tab:fb}
\begin{center}
\begin{tabular}{ccccccc}
\hline
\multicolumn{1}{c}{\bf Classifier/}
&\multicolumn{1}{c}{\bf No Attack}
&\multicolumn{1}{c}{\bf No Defense}
&\multicolumn{1}{c}{\bf LQ-VAE}
&\multicolumn{1}{c}{\bf Defense-GAN}
&\multicolumn{1}{c}{\bf Madry}
&\multicolumn{1}{c}{\bf Adv Tr}\\
{\bf Substitute}&&&&&&
\\ \hline 
A/B & 92.76 & 29.14 & 77.74 & 74.41 & 60.11 & 48.27 \\
A/C & 92.76 & 35.44 & 77.11 & 74.11 & 62.58 & 57.53 \\
B/A & 91.17 & 67.82 & 81.33 & 77.97 & 80.71 & 76.61 \\
B/C & 91.17 & 45.55 & 78.83 & 74.5  & 69.19 & 64.05 \\
C/A & 89.06 & 79.11 & 82.12 & 78.82 & 80.99 & 81.84 \\
C/B & 89.06 & 47.26 & 80.76 & 76.6  & 67.46 & 59.64 \\
\hline
{\bf Average}& 91.00 & 50.72 & \bf 79.65 & 76.07 & 70.17 & 64.66\\
\hline 
\end{tabular}
\caption{Classification accuracy of the FMNIST classifier on Deepfool black box attack images generated using substitute model.}
\end{center}

\end{table*}
\begin{table*}[!t]
\label{celabb}
\begin{center}
\begin{tabular}{ccccccc}
\hline
\multicolumn{1}{c}{\bf Classifier/}
&\multicolumn{1}{c}{\bf No Attack}
&\multicolumn{1}{c}{\bf No Defense}
&\multicolumn{1}{c}{\bf LQ-VAE}
&\multicolumn{1}{c}{\bf Defense-GAN}
&\multicolumn{1}{c}{\bf Madry}
&\multicolumn{1}{c}{\bf Adv Tr}\\
{\bf Substitute}&&&&&&
\\ \hline 
A/B & 96.34 & 39.53 & 86.01 & 84.70 & 85.41 & 94.13 \\
A/C & 96.34 & 37.59 & 80.10 & 78.11 & 64.72 & 54.22 \\
B/A & 96.60 & 49.21 & 85.67 & 86.19 & 82.55 & 68.11 \\
B/C & 96.60 & 52.52 & 79.98 & 79.91 & 76.31 & 62.53 \\
C/A & 95.02 & 82.87 & 85.90 & 86.29 & 89.79 & 86.75 \\
C/B & 95.02 & 83.26 & 85.20 & 86.17 & 89.91 & 88.40 \\
\hline
{\bf Average}& 95.99 & 57.50 & \bf 83.81 & 83.56 & 81.45 & 75.69\\
\hline 
\end{tabular}
\caption{classification accuracy of the CelebA classifier on CW black box attack images generated using substitute model}
\end{center}

\end{table*}
\begin{table}[!t]
\begin{center}
\begin{tabular}{ccc}
\hline
\multicolumn{1}{c}{\bf Dataset}
&\multicolumn{1}{c}{\bf LQ-VAE}
&\multicolumn{1}{c}{\bf Defense-GAN}
\\ \hline
MNIST 
& 83.70 & 55.17\\
FMNIST 
& 57.41 & 39.41\\
\hline
\end{tabular}
\caption{Classification accuracy of end-to-end whitebox FGSM attack on LQ-VAE - classifier combination using BPDA.}
\label{tab:BPDA}
\end{center}

\end{table}
%


\subsection{Results on black box attacks} 
For black box experiments, we consider one attack each on a dataset as a representative set. Specifically, FGSM for MNIST, Deepfool for FMINST and CW for CelebA are considered with six-pairs of classifiers used for attacking and substitute. A similar trend is observed with the black box attacks as well, as seen in Tables 4, 5 and 6. Madry retraining  performs the best on the FGSM attack because it is trained on a superset of first-order methods of which FGSM is a subset. However, the performance of LQ-VAE is consistent irrespective of the classifier pairs across all cases and is closely comparable (or better) to the best case. On Deepfool and CW, LQ-VAE outperforms the others in most of the cases. We hypothesize that this behaviour of LQ-VAE comes from the Lipschitz constraining, by which it becomes a strong defense when the attack alters fewer pixels of the input image yet changing the class, as in the case of CW and Deepfool, unlike in FGSM.  In summary, the proposed method is invariant to white or black box attack types.

\section{Discussions and Conclusions}

LQ-VAE and Defense-GAN fall into the same category of defense mechanisms in the sense that they both capitalize on the expressive capacity of generative models. Further, both of the models generalize better on the adaptive or unseen attacks \cite{athalye2018obfuscated} since both of them neither need access to the classifier nor train on a certain type of adversaries. However, LQ-VAE offers several advantages over Defense-GAN such as - (i) LQ-VAE does not involve a run-time search on the latent space unlike Defense-GAN which makes it orders of magnitude faster and independent of latent search parameters. Rather in LQ-VAE, the search in the latent space is implicitly done  by effective encoding, quantization and decoding of the latent space. (ii) training a VAE is known to be easier and faster, yielding a better data likelihood than a GAN which is known to be difficult to be trained, especially on color datasets such as CelebA, (iii) LQ-VAE has a latent encoding followed by the re-mapping of the latent space to the data space which makes it invariant to attack types while Defense-GAN is shown to degrade in the case of black box attacks, (iv) as argued in \cite{athalye2018obfuscated}, Defense-GAN can be attacked too by a method called the Backward Pass Differentiable Approximation (BPDA), in which case its defense on MNIST is reported at 55\% \cite{athalye2018obfuscated}. When the same technique is used to attack LQ-VAE, we get much better accuracy of 83\% on the same task which can be ascribed to the use of latent space constraining and quantization. In summary, we proposed a technique called LQ-VAE as a filter for the adversarial examples using a constrained projection on to a quantized latent space followed by data reconstruction. It serves like a `black-box defense' in the sense that it can be used to defend any attack and with any classifier. In principle, LQ-VAE can be re-trained using adversaries too, in which case the performance is observed to improve. For instance, it is observed that if one retrains LQ-VAE using Madry adversaries, its performance is enhanced by 5-10\% on FGSM attacks. Future directions include exploration of the latent prior $p_\theta$ beyond a standard Normal distribution, studying the effect of different types of quantization other than a simple binary quantization and using LQ-VAE as an adversarial detector. We provide the code\footnote{The code used in this paper can be accessed at \url{https://github.com/mayank31398/lqvae}} for further research.

\bibliography{references}
\bibliographystyle{aaai}

\appendix
\section{Appendix}

\subsection{Proof for Theorem 1}
Let $\hat{\rvz}$ and $\hat{\rvz}_{adv}$ be the latent encoding of $\rvx$ and $\rvx_{adv}$ respectively. Let $\rvz = H(\hat{\rvz})$ and $\rvz_{adv} = H(\hat{\rvz}_{adv})$ be the corresponding quantizations with $\eta$ being the quantization threshold. Then the probability of a particular bit being flipped is given by:
\begin{equation}
    p = \underbrace{P(\rvz = -1, \rvz_{adv} = 1)}_A + \underbrace{P(\rvz = 1, \rvz_{adv} = -1)}_B
\end{equation}

\begin{equation}
    \begin{aligned}
        A & = P(\hat{\rvz} < -\eta, -\eta < \hat{\rvz}_{adv} < \eta)
        \\
        & + P(\hat{\rvz} > \eta, -\eta < \hat{\rvz}_{adv} < \eta)
        \\
        \\
        & = P(\hat{\rvz} < -\eta, -\eta - \delta_z < \hat{\rvz} < \eta - \delta_z)
        \\
        & + P(\hat{\rvz} > \eta, -\eta - \delta_z < \hat{\rvz} < \eta - \delta_z)
        \\
        \\
        & = \left\{
            \begin{array}{ll}
                \int_{-\eta - \delta_z}^{-\eta} p_{\hat{\rvz}} (\hat{\rvz}) d\hat{\rvz} & \text{if } \delta_z \ge 0
                \\
                \\
                \int_{\eta}^{\eta - \delta_z} p_{\hat{\rvz}} (\hat{\rvz}) d\hat{\rvz} & \text{otherwise}
            \end{array}
            \right.
        \\
        & = \int_{\eta}^{\eta + |\delta_z|} p_{\hat{\rvz}} (\hat{\rvz}) d\hat{\rvz}
    \end{aligned}
\end{equation}
which is obtained by using $p_{\hat{\rvz}} (\hat{\rvz}) = \mathcal{N}(0, 1)$ which is an even function. Similarly,
\begin{equation}
    B = \int_{\eta - |\delta_z|}^\eta p_{\hat{\rvz}} (\hat{\rvz}) d\hat{\rvz}
\end{equation}
\\
Adding A and B we have,
\begin{equation}
    p = \int_{\eta - |\delta_z|}^{\eta + |\delta_z|} p_{\hat{\rvz}} (\hat{\rvz}) d\hat{\rvz}
\end{equation}

To give equal probability to $\rvz$ taking the values -1 or 1 we have the following constraint:
\begin{equation}
    P(\rvz = 1) = \int_{-\eta}^{\eta} p_{\hat{\rvz}} (\hat{\rvz}) d\hat{\rvz} = 0.5
\end{equation}
Solving this for $p_{\hat{\rvz}} (\hat{\rvz}) = \mathcal{N}(0, 1)$, we have $\eta = 0.67449$. Using this value of $\eta$ and $|\delta_z| = K|\delta|$ we get $|\delta_z| = 0.03$ where $K = 0.1$ is the Lipschitz constant and $|\delta| = 0.3$ is the perturbation in the input image. Thus, the probability of a bit being flipped is found to be $p = 0.01906$ which is a significantly low number. For a $M$ dimensional latent space, the probability of $k$ bit flips is ${M \choose k} p^k (1-p)^{M-k}$ which is found to be $1.177 \times 10^{-3}$ for $k = 6$ ($\approx 10\%$) bit flips for a $M = 64$ dimensional latent space.

\subsection{Architectures of the classifier and substitute networks}
The following table shows the neural network architectures used throughout the paper for classifier and substitute models.
Terminology used:

\begin{itemize}[noitemsep]
    \item Conv($m, f \times f, s$) denotes a convolutional layer with m filters of size $f \times f$ and stride $s$
    \item ReLu is the Rectified Linear Unit Activation
    \item LeakyReLu($\alpha$) is the Leaky Rectified Linear Unit Activation with parameter $\alpha$
    \item Dropout($p$) is a dropout layer with probability $p$
    \item FC($m$) denotes a fully connected layer with m neuron units
    \item ConvT($m, f\times f, s$) denotes a deconvolutional layer with $m$ filters of size $f \times f$and stride $s$
\end{itemize}

\begin{table}[H]
    \centering
    \begin{tabular}{c|c|c}
        A & B & C
        \\
        \hline
        Conv(64, $5\times5$, 1) & Dropout(0.2) & FC(200)
        \\
        ReLu & Conv(64, $8\times8$, 2) & ReLu
        \\
        Conv(64, $5\times5$, 2) & ReLu & Dropout(0.5)
        \\
        ReLu & Conv(128, $6\times6$, 2) & FC(200)
        \\
        Dropout(0.25) & ReLu & ReLu
        \\
        FC(128) & Conv(128, $5\times5$, 1) & FC($*$)
        \\
        ReLu & ReLu & Softmax
        \\
        Dropout(0.5) & Dropout(0.5) &
        \\
        FC($*$) & FC($*$) &
        \\
        Softmax & Softmax &
    \end{tabular}
\end{table}
{\centering Final Fully connected layer has 10 units for MNIST and FMNIST and 2 units for CelebA dataset.\par}

\subsection{LQ-VAE architecture}
\begin{table}[H]
    \centering
    \begin{tabular}{c |c }
        Encoder & Decoder1\textbackslash Decoder2
        \\
        \hline
        Conv(64, $3\times3$, 1)  & FC(1024)
        \\
        LeakyReLu(0.2) & LeakyReLu(0.2)
        \\
        Conv(64, $3\times3$, 2)  & FC(6272)
        \\
        LeakyReLu(0.2) & LeakyReLu(0.2)
        \\
        Conv(128, $3\times3$, 2)  & ConvT(128,$3\times3$, 1)
        \\
        LeakyReLu(0.2) & LeakyReLu(0.2)
        \\
        Conv(128, $3\times3$, 1)  & ConvT(128,$3\times3$, 2)
        \\
        LeakyReLu(0.2) & LeakyReLu(0.2)
        \\
        FC(1024)   & ConvT(64,$3\times3$, 2)
        \\
        LeakyReLu(0.2) & LeakyReLu(0.2)
        \\
        FC(64), FC(64)  & ConvT(1,$3\times3$, 1)
    \end{tabular}
\caption{The encoder and decoder of LQ-VAE used in the experiments on MNIST and Fashion-MNIST}
\end{table}

\begin{table}[H]
    \centering
    \begin{tabular}{c |c }
        Encoder & Decoder1\textbackslash Decoder2
        \\
        \hline
        Conv(64, $3\times3$, 1)  & FC(1024)
        \\
        LeakyReLu(0.2) & LeakyReLu(0.2)
        \\
        Conv(64, $3\times3$, 2)  & FC(4096)
        \\
        LeakyReLu(0.2) & LeakyReLu(0.2)
        \\
        Conv(128, $3\times3$, 2)  & ConvT(128,$3\times3$, 2)
        \\
        LeakyReLu(0.2) & LeakyReLu(0.2)
        \\
        Conv(128, $3\times3$, 2)  & ConvT(128,$3\times3$, 2)
        \\
        LeakyReLu(0.2) & LeakyReLu(0.2)
        \\
        Conv(256, $3\times3$, 2)  & ConvT(64,$3\times3$, 2)
        \\
        LeakyReLu(0.2) & LeakyReLu(0.2)
        \\
        FC(1024)   & ConvT(64,$3\times3$, 2)
        \\
        LeakyReLu(0.2) & LeakyReLu(0.2)
        \\
        FC(64), FC(64)  & ConvT(3,$3\times3$, 1)
    \end{tabular}
\caption{The encoder and decoder of LQ-VAE used in the experiments on CelebA}
\end{table}

\end{document}